\documentclass[conference]{IEEEtran}
\IEEEoverridecommandlockouts
\usepackage{cite}
\usepackage{amsmath,amssymb,amsfonts}
\usepackage{algorithmic}
\usepackage{graphicx}
\usepackage{textcomp}
\usepackage{xcolor}
\usepackage{threeparttable}
\def\BibTeX{{\rm B\kern-.05em{\sc i\kern-.025em b}\kern-.08em
    T\kern-.1667em\lower.7ex\hbox{E}\kern-.125emX}}

\makeatletter
\newcommand{\linebreakand}{%
  \end{@IEEEauthorhalign}
  \hfill\mbox{}\par
  \mbox{}\hfill\begin{@IEEEauthorhalign}
}
\makeatother
    
\begin{document}

\title{LB-KBQA:Large-language-model and BERT based Knowledge-Based Question and Answering System\\
}
\author{\IEEEauthorblockN{1\textsuperscript{st} Yan Zhao}
\IEEEauthorblockA{\textit{XJTLU} \\
Suzhou, China \\
gzxzhaoyan@gmail.com}

\and
\IEEEauthorblockN{2\textsuperscript{nd} Zhongyun Li}
\IEEEauthorblockA{\textit{University of Liverpool} \\
Liverpool, UK \\
lizy10@liverpool.ac.uk}

\and
\IEEEauthorblockN{3\textsuperscript{rd} Yushan Pan}
\IEEEauthorblockA{\textit{dept. School of Advanced Technology} \\
\textit{XJTLU}\\
Suzhou, China \\
yushan.pan@xjtlu.edu.cn}

\linebreakand

\and

\IEEEauthorblockN{4\textsuperscript{th} Jiaxing Wang}
\IEEEauthorblockA{\textit{University of Liverpool} \\
Liverpool, UK \\
J.Wang214@liverpool.ac.uk}
\and

\IEEEauthorblockN{5\textsuperscript{th} Yihong Wang}
\IEEEauthorblockA{\textit{dept. School of Advanced Technology} \\
\textit{XJTLU}\\
Suzhou, China \\
yihong.wang@xjtlu.edu.cn}

}
%\linebreakand
\maketitle

\begin{abstract}
Generative Artificial Intelligence (AI), because of its emergent abilities, has empowered various fields, one typical of which is large language models (LLMs). One of the typical application fields of Generative AI is large language models (LLMs), and the natural language understanding capability of LLM is dramatically improved when compared with conventional AI-based methods. The natural language understanding capability has always been a barrier to the intent recognition performance of the Knowledge-Based-Question-and-Answer (KBQA) system, which arises from linguistic diversity and the newly appeared intent. Conventional AI-based methods for intent recognition can be divided into semantic parsing-based and model-based approaches. However, both of the methods suffer from limited resources in intent recognition. To address this issue, we propose a novel KBQA system based on a Large Language Model(LLM) and BERT (LB-KBQA). With the help of generative AI, our proposed method could detect newly appeared intent and acquire new knowledge. In experiments on financial domain question answering, our model has demonstrated superior effectiveness.
\end{abstract}

\begin{IEEEkeywords}
Generative AI, KBQA, LLM
\end{IEEEkeywords}

\section{Introduction}
Recently, Generative Artificial Intelligence (AI) has gained much attention from both academia and industry as the emergent abilities of generative AI have great potential to empower various fields. Generative AI covers a wide range of applications, the most typical of which is large language models (LLMs)\cite{b1}. A famous example is the GPT series\cite{b2}, which has made noteworthy advancements in this domain\cite{b3}. The GPT series models represent typical examples of large-scale language models. Scholars have observed that as the parameter count of GPT language models grows, their natural language understanding capabilities become increasingly powerful. For instance, GPT-3 boasts 170 billion parameters\cite{b4}. This natural language understanding capability holds significant promise in both business and society and offers considerable potential in various domains\cite{b5}.

The ability of natural language understanding is the most limiting factor in the field of question-answering (QA) systems. The QA systems are utilized in open-world scenarios, which often result in issues of questions and answers (QA) mismatching problems due to linguistic diversity \cite{b6}. The open-world scenario for the QA system usually means the training set is typically unable to encompass all possible user question inputs. The different expressions of the questions from users and the linguistic diversity usually lead to an unseen class problem \cite{b7}, and this problem will further cause the QA mismatching problem and undermine the QA matching accuracy. Specifically, the negative impact caused by the unseen classes will be a disaster in the field of QA tasks with a fixed knowledge base, so-called Knowledge-based Question and Answer (KBQA) \cite{b8}, as the mismatching problems will not just hurt the accuracy but might directly lead to the task failure.

With the rapid development of large-scale knowledge bases, the open-domain Knowledge-based Question and Answer (KBQA) has emerged. KBQA system is to understand natural language questions and answers based on external knowledge\cite{b9}. The field of KBQA has gained significant attention from both academia and industry in recent years. The open-source knowledge bases have enabled KBQAs to provide accurate answers across various domains. For example, the broad knowledge graph called Knowledge Vault (KV)\cite{b10} was integrated into Google's search engine for practical purposes and resulted in significant improvement in the user experience of the search engines. The basic idea of the KBQA is to find a mapping from the natural language question of users to the answers in the knowledge base. Additionally, the first step of the KBQA is to understand the question from the semantic level, and the questions are usually mapped to the intent of the users. The intents can be considered as fixed classes, and in essence, the intent recognition task will encounter the problem of \textbf{\emph{unseen classes}}, or to be more precise, the \textbf{\emph{unseen intents}}. Therefore, the problem of the \textbf{\emph{unseen intents}} will directly cause the failure of the KBQA system.

This mapping is a difficult task in that it might introduce errors from linguistic diversity as there might be various ways to describe the same subject. Optimizing these systems requires correctly analyzing human language and providing appropriate responses. For the user intent natural language understanding (NLU), the intent recognition of KBQA can be divided into two technical paths: 1) the rule-based method and 2) the model-based method. Recently, semantic parsing has been one of the rule-based methods and has been the mainstream method in question-and-answer systems\cite{b11}. However, the rule-based method ignores high-dimensional semantic information and struggles to handle complex queries that involve multiple entities. Additionally, the model-based method can capture the semantic information by the conventional AI (e.g., BB\_KBQA\cite{b12}). Still, model-based approaches frequently encounter two challenges: firstly, they often struggle to provide precise answers for language data beyond the pre-training dataset; secondly, the expense of models acquiring new knowledge is comparatively high.

In this case, the Intent Recognition Task might suffer from severe failure caused by \textbf{\emph{unseen intents}} at the very beginning. The \textbf{\emph{unseen intents}} are very common in the application field of KBQA and can be considered into two categories, 1) \textbf{\emph{unseen intents}} caused by linguistic diversity (e.g., different question representations of the same intent), and 2) the intents that have never been considered by the knowledge base. 

To tackle the limitation of the \textbf{\emph{unseen intents}}, we creatively introduce the large language model. Recently, the development of large-scale language models has represented a significant breakthrough in natural language processing\cite{b2,b13}. These models are artificial intelligence systems designed to generate text that is highly similar to human-generated text. They achieve this by processing vast quantities of textual data, enabling them to identify and learn various linguistic patterns and habits exhibited by users\cite{b14}. Moreover, large language models can quickly acquire new knowledge through prompt learning, making them highly adaptable and capable of improving their performance in response to user input and exposure to novel information\cite{b13}. The impressive performance of large language models has the potential to address the \textbf{\emph{unseen intents}} problem.

To solve the problem of the \textbf{\emph{unseen intents}} in the field of KBQA, we introduced our system named LB-KBQA, which could tackle intentions misunderstanding issues. Our knowledge base consists of two main components: an intent library that utilizes vector representation and a query library that utilizes the knowledge graph. The system is composed of five main parts, each with its specific function. Firstly, the language preprocessing module removes irrelevant symbols from the input text. The intention recognition module, which is the second part, comprises a rule-based intention recognition model, a high-dimensional semantic representation module based on BERT, and an \textbf{\emph{unseen intents}} processing module based on pre-trained language models. By fusing these three models, the system effectively addresses the challenge of intention recognition failure caused by language diversity, which is explained in detail in the system design section. The third part is the answer generation module, in which the user's question and the matched answer are sent to LLM (large language models), and more readable answers are generated through prompt learning. To overcome the issue of intention recognition failure stemming from limitations of the knowledge base, we developed an adaptive learning module as the fourth part. The module gradually obtains the user's true intention through interaction between LLM and the user and updates it to the intent library. Finally, we designed a query library extension module to facilitate batch updates of the knowledge graph for users.

The contribution can be summarized as follows.

\begin{itemize}
\item[$\bullet$]The proposed method identifies the unseen intent for a specific-domain KBQA system. The unseen intents are divided into two parts, the diverse representation of intents and the newly appeared intents, and both of the unseen intents can be captured by the proposed method.
\end{itemize}

\begin{itemize}
\item[$\bullet$]We provide a solution for building a KBQA system in the financial domain based on Generative AI.

\end{itemize}

\begin{figure*}[t]
\centering
\includegraphics[width=\textwidth]{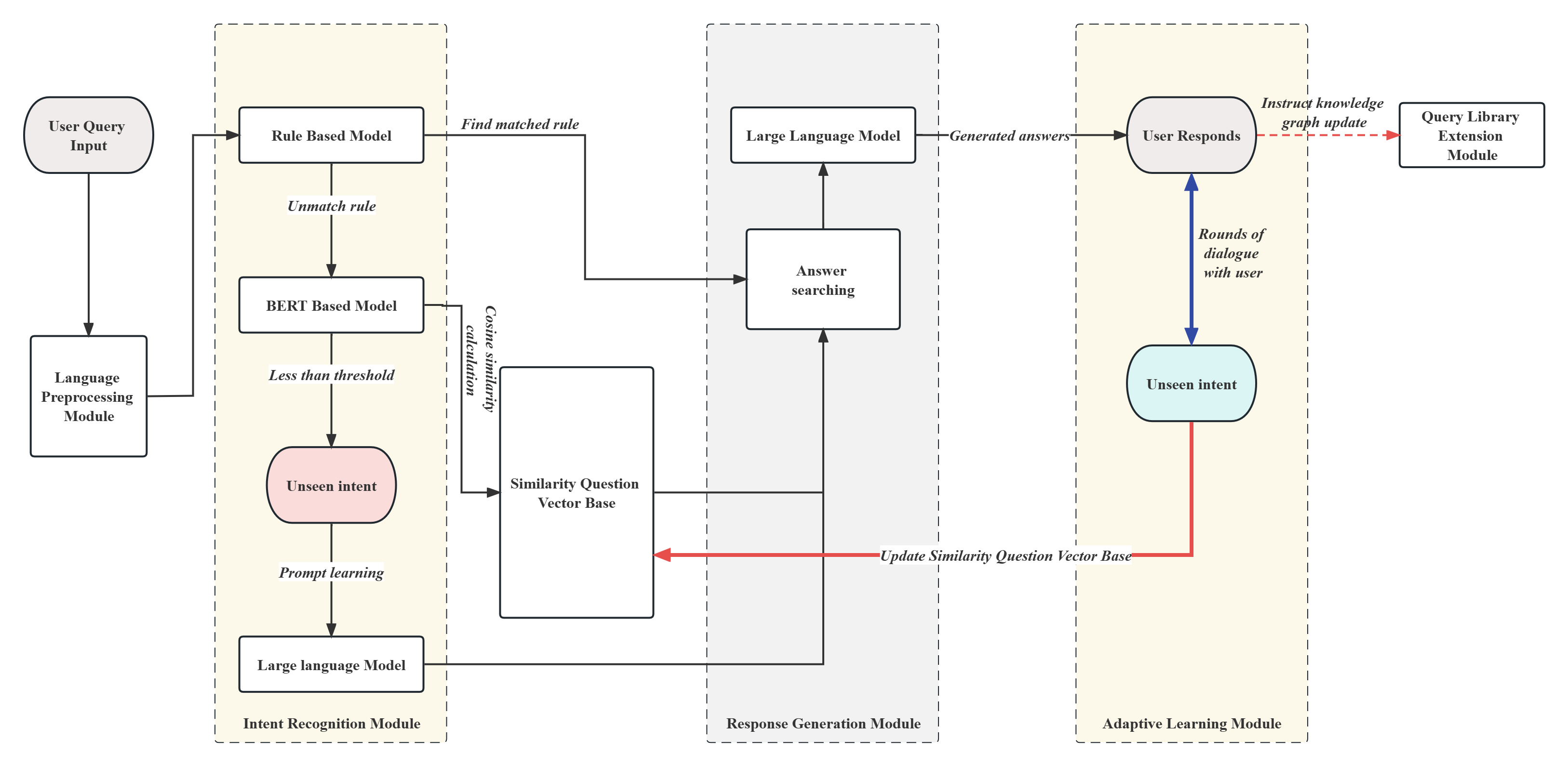}
\caption{The LB-KBQA system is comprised of five distinct components: 1) the language preprocessing module, 2) the intent recognition module, 3) the response generation module, 4)  the adaptive learning module, and 5) the query library extension module.}
\label{fig:example}
\end{figure*}

\section{System Design}

Figure 1 illustrates the system design, which comprises five main parts: 1) the language preprocessing module, 2) the intent recognition module, 3) the response generation module, 4)  the adaptive learning module, and 5) the query library extension module. In this section, we will provide details about each of these components.

\subsection{Language Preprocessing Module}

The language preprocessing module is responsible for removing stop words, punctuation, and special symbols from the input text\cite{b15}. This process improves search efficiency and accuracy, as stop words and special symbols such as emojis do not provide any meaningful information. For example, words like "the," "is," and "in" are removed in English.

\subsection{Intent Recognition Module}

The intent recognition module identifies the key entities in the input request. For example, if the request is related to time, the module selects key entities such as "year," "month," and "day." Accurate intent recognition is critical for providing accurate answers in a KBQA system.

Figure 1 shows the details of the intent recognition process. Firstly, the clean text will be compared with the question rule, which is a domain-specific semantic analysis model containing semantic rules. The rules are used to match the text with the intent label. The rule-based model has the advantage of being fast and scenario-oriented, but it cannot handle queries with complex semantics. If the rule-based model fails to make a prediction, the query is passed to the question embedding component, which extracts high-dimensional semantic information using the BERT model. The input query is converted to a context vector for further similarity calculation. Our implementation uses a similarity question vector base to store the questions in tuple type, which includes the intent label and context vector. The cosine similarity method is used to measure the similarity between the context vector and the corresponding vector. The intention label that refers to the corresponding vector which is most similar will be selected\cite{b16}.

However, utilizing the BERT model for question embedding may not entirely correspond to the diverse nature of natural language, as it cannot account for the potential variability in questioning habits and speaking styles. This variability in language usage may result in semantic errors in question embedding, which can subsequently impact the vector representation. In order to address this issue, a large language model is employed as a fallback approach. Initially, a threshold for cosine similarity is established. If the question vector represented by BERT is not deemed similar enough to any of the vectors in the Similarity Question Vector Base, as determined by this threshold, then the large model is implemented for intent determination. In the intent generation process, an in-context learning method is applied for prompt learning. The logical reasoning process for intent determination is regarded as input for the large model, along with the user's query. Consequently, the large model can generate user intent based on the user's input. Meanwhile, the generated intent labels will be subsequently updated in the intent library.

\subsection{Response Generation Module}

The target of the Response Generation Module is to provide a readable answer according to user intents. It has three steps. First, the Jieba package is used to perform named entity recognition (NER) tasks. Then, the selected entities are converted into a query statement. As the knowledge graph is built using neo4j, the query language cipher is used to query the graph. The query statement includes a symbol, "XX," which is replaced with real entities. Finally, we prompt the Large Language Model with user query and matched knowledge to generate a readable answer. 

\subsection{Adopted Learning Module}

The purpose of the Adaptive Learning Module is to enhance the understanding capability of unknown intents for the question-answering system. As illustrated in Figure 1, the working mechanism of this module is as follows: if the user expresses dissatisfaction with the answer provided by the Knowledge-Based Question Answering system, the large language model will engage in a multi-turn dialogue with the user, based on preset prompts, to confirm 1) whether the user's dissatisfaction is due to intent recognition issues, and 2) what the user believes the correct intent is. Once the user provides the correct intent, the intent label is updated together with the question vector in the Similarity Question Vector Base. It should be noted that this module does not  consider cases where user dissatisfaction is caused by non-intent recognition issues. By designing the adaptive learning module, the system can achieve rapid adaptation to unknown intents.

\subsection{Query Library Extension Module}

Users can use the py2neo Python package to extend the existing knowledge graph with structured datasets. For unstructured datasets, semantic analysis toolkits are recommended to extract the entities and relationships based on semantic parsing. The query library extension module enables users to build a domain-specific knowledge graph and integrate it into the question-and-answer system.

\section{Case Study}

\subsection{Experiment Setting}

In typical question answering (QA), the confusion matrix is a commonly used evaluation factor. Previous studies, such as QASE(Question Answering with Subgraph Embeddings)\cite{b17} and MCCNN(Multi-Column Convolutional Neural Networks)\cite{b18}, have introduced the confusion matrix to evaluate the results of their methods.  The previous research is based on general datasets, such as DBpedia and Freebase, which involves cross-domain datasets, and there exists the assumption of failure question matching. However, our QA system is designed specifically within a domain-specific knowledge base and does not take failure question matching assumption into consideration. Furthermore, our system is designed to address the problem of identifying unseen intent while disregarding answer failures caused by insufficient data. As a result, we have made a fundamental assumption that all questions can find corresponding answers within the domain knowledge base. Based on this, we still utilize the classic QA evaluation standard\cite{b19}, but we contend that accuracy alone is sufficient to assess the efficacy of the system.

To create our system, we utilized the Tushare financial dataset (http://tushare.org/), an open-source, python-friendly financial data package that offers users fast, clean, and diverse data. 

Using the Tushare financial dataset, we created our knowledge graph and a test set containing 100 samples, comprising 50 simple relation questions and 50 complex relation questions. A simple relation question refers to a question whose answer can be found within a triplet of SPOs. For instance, if we ask, "Where is the headquarters of Wanke company located?" the answer "Shenzhen" is stored in the simple triplet as $<wanke - located - Shenzhen>$. In contrast, a complex relation question involves the system needing to search more than one SPO triplet to generate a final answer. An example of a complex relation question is "Please name the five most popular investment companies."

\subsection{Experiment Result}

We conducted ablation experiments to compare the impact of different model settings on the performance of our system. The abbreviation w/o, denoting the removal of a particular part from the model, is used in Table \ref{tab:Results} to illustrate the experimental groups. In this study, accuracy was adopted as the primary evaluation metric, calculated by dividing the number of correct answers by the total number of questions. Our experimental results show that the complete system for the question-answering task achieved an accuracy of 0.90. It should be emphasized that the system's ability to correct incorrect answers through adaptive learning is regarded as equivalent to providing the correct answers.

In order to evaluate the system's recognition performance for unseen intents, we conducted separate tests for various model settings. Our evaluation results demonstrate that the BERT-based question representation model and the similar question vector library are crucial for discovering unseen intents. The experimental results show that the accuracy decreased by 0.3 after removing the BERT module. Furthermore, the incorporation of LLM and Adaptive Learning Module can effectively address the diversity of language and improve the system's ability to recognize unseen intents. In contrast, the rule-based model has a limited impact on the system's recognition of unseen intents.

\begin{table}[!ht]
    \renewcommand{\arraystretch}{1.5}
    \centering
    \caption{Evaluation results of different settings on the test split of the financial dataset.}
    \tabcolsep=0.7cm
    \begin{threeparttable}          %这行要添加
    
    \begin{tabular}{ll}
    \hline
        \textbf{Setting} & \textbf{Accuracy} \\ \hline
        \textbf{All} & \textbf{0.90} \\ 
        \textbf{w/o Rule Based Model}\tnote{1} & 0.88 \\ 
        \textbf{w/o Bert Based Model}\tnote{2} & 0.60 \\ 
        \textbf{w/o LLM}\tnote{3} & 0.8 \\ 
        \textbf{w/o Adaptive Learning Module}\tnote{4} & 0.85 \\ \hline
    \end{tabular}
    \label{tab:Results}
        \begin{tablenotes}    %这行要添加， 从这开始
            \footnotesize               %这行要添加
            \item[1] w/o Rule Based Model: without using the Rule Based Model in the intent recognition module.
            \item[2] w/o Bert Based Model: without using the Bert Based Model in the intent recognition module.  
            \item[3] w/o LLM: without using the LLM in the intent recognition module. 
            \item[4] w/o Adaptive Learning Module: without using the Adaptive Learning Module
      \end{tablenotes}            %这行要添加
    \end{threeparttable}       %这行要添加，到这里结束
\end{table}

\section{Conclusion}
We conducted ablation experiments to compare the impact of different model settings on the performance of our system. The abbreviation w/o, denoting the removal of a particular part from the model, is used in Table 1 to illustrate the experimental groups. In this study, accuracy was adopted as the primary evaluation metric, calculated by dividing the number of correct answers by the total number of questions. Our experimental results show that the complete system for the question-answering task achieved an accuracy of 0.90. It should be emphasized that the system's ability to correct incorrect answers through adaptive learning is regarded as equivalent to providing the correct answers.

In order to evaluate the system's recognition performance for unseen intents, we conducted separate tests for various model settings. Our evaluation results demonstrate that the BERT-based question representation model and the similar question vector library are crucial for discovering unseen intents. The experimental results show that the accuracy decreased by 0.3 after removing the BERT module. Furthermore, the incorporation of Generative AI and Adaptive Learning Modules can effectively address the diversity of language and improve the system's ability to recognize unseen intents. In contrast, the rule-based model has a limited impact on the system's recognition of unseen intents. 

Our proposed method provides evidence in the KBQA field that the natural language understanding capability of Generative AI effectively helps the traditional AI methods to tackle the barrier of linguistic diversity. Additionally, in an industrial context, we provide methodological contributions to solving the problem of unseen classes in the field of KBQA. 
Furthermore, at the application level, we present a Generative AI-based solution for constructing KBQA systems in the financial domain. Through ablation experiments, we verify the effectiveness of Generative AI in enhancing the natural language understanding capability of traditional AI.

%% The file named.bst is a bibliography style file for BibTeX 0.99c

$\bibliographystyle{named}
$\bibliography{B}

\begin{thebibliography}{00}


\bibitem{b1} S. Mohamadi, G. Mujtaba, N. Le, G. Doretto, and D. A. Adjeroh, “ChatGPT in the Age of Generative AI and Large Language Models: A Concise Survey,” 2023, doi: 10.48550/ARXIV.2307.04251.
\bibitem{b2} C. Zhou et al., “A Comprehensive Survey on Pretrained Foundation Models: A History from BERT to ChatGPT,” 2023, doi: 10.48550/ARXIV.2302.09419.
\bibitem{b3} A. Bandi, P. V. S. R. Adapa, and Y. E. V. P. K. Kuchi, “The Power of Generative AI: A Review of Requirements, Models, Input–Output Formats, Evaluation Metrics, and Challenges,” Future Internet, vol. 15, no. 8, p. 260, Jul. 2023, doi: 10.3390/fi15080260.
\bibitem{b4} Y. Chang et al., “A Survey on Evaluation of Large Language Models.” arXiv, Oct. 17, 2023. Accessed: Oct. 21, 2023. [Online]. Available: http://arxiv.org/abs/2307.03109
\bibitem{b5} A. Carlson, J. Betteridge, B. Kisiel, B. Settles, E. Hruschka, and T. Mitchell, “Toward an Architecture for Never-Ending Language Learning,” AAAI, vol. 24, no. 1, pp. 1306–1313, Jul. 2010, doi: 10.1609/aaai.v24i1.7519.
\bibitem{b6} C. Deng, G. Zeng, Z. Cai, and X. Xiao, “A survey of knowledge based question answering with deep learning,” Journal of Artificial Intelligence, vol. 2, no. 4, p. 157, 2020.
\bibitem{b7} J. Parmar, S. Chouhan, V. Raychoudhury, and S. Rathore, “Open-world Machine Learning: Applications, Challenges, and Opportunities,” ACM Comput. Surv., vol. 55, no. 10, pp. 1–37, Oct. 2023, doi: 10.1145/3561381.
\bibitem{b8} W. Wang, Y. Xiao, and W. Cui, “KBQA: An Online Template Based Question Answering System over Freebase,” in Proceedings of the Twenty-Fifth International Joint Conference on Artificial Intelligence (IJCAI-16), New York, NY, USA, 2016, pp. 9–15.
\bibitem{b9} J. Yuan et al., “Constructing biomedical domain-specific knowledge graph with minimum supervision,” Knowl Inf Syst, vol. 62, no. 1, pp. 317–336, Jan. 2020, doi: 10.1007/s10115-019-01351-4.
\bibitem{b10} X. Dong et al., “Knowledge vault: a web-scale approach to probabilistic knowledge fusion,” in Proceedings of the 20th ACM SIGKDD international conference on Knowledge discovery and data mining, New York New York USA: ACM, Aug. 2014, pp. 601–610. doi: 10.1145/2623330.2623623.
\bibitem{b11} J. Berant, A. Chou, R. Frostig, and P. Liang, “Semantic parsing on freebase from question-answer pairs,” in Proceedings of the 2013 conference on empirical methods in natural language processing, 2013, pp. 1533–1544.

\bibitem{b12} A. Liu, Z. Huang, H. Lu, X. Wang, and C. Yuan, “BB-KBQA: BERT-Based Knowledge Base Question Answering,” in Chinese Computational Linguistics, vol. 11856, M. Sun, X. Huang, H. Ji, Z. Liu, and Y. Liu, Eds., in Lecture Notes in Computer Science, vol. 11856. , Cham: Springer International Publishing, 2019, pp. 81–92. doi: 10.1007/978-3-030-32381-3\_7.
\bibitem{b13} P. Liu, W. Yuan, J. Fu, Z. Jiang, H. Hayashi, and G. Neubig, “Pre-train, Prompt, and Predict: A Systematic Survey of Prompting Methods in Natural Language Processing,” ACM Comput. Surv., vol. 55, no. 9, pp. 1–35, Sep. 2023, doi: 10.1145/3560815.
\bibitem{b14} R. Balestriero et al., “A Cookbook of Self-Supervised Learning,” 2023, doi: 10.48550/ARXIV.2304.12210.
\bibitem{b15} L. Zheng, Z. He, and S. He, “A novel probabilistic graphic model to detect product defects from social media data,” Decision Support Systems, vol. 137, p. 113369, Oct. 2020, doi: 10.1016/j.dss.2020.113369.
\bibitem{b16} L. Zou, R. Huang, H. Wang, J. X. Yu, W. He, and D. Zhao, “Natural language question answering over RDF: a graph data driven approach,” in Proceedings of the 2014 ACM SIGMOD international conference on Management of data, 2014, pp. 313–324.
\bibitem{b17} A. Bordes, S. Chopra, and J. Weston, “Question answering with subgraph embeddings,” arXiv preprint arXiv:1406.3676, 2014.
\bibitem{b18} L. Dong, F. Wei, M. Zhou, and K. Xu, “Question answering over freebase with multi-column convolutional neural networks,” in Proceedings of the 53rd Annual Meeting of the Association for Computational Linguistics and the 7th International Joint Conference on Natural Language Processing (Volume 1: Long Papers), 2015, pp. 260–269.
\bibitem{b19} X. Yao and B. Van Durme, “Information extraction over structured data: Question answering with freebase,” in Proceedings of the 52nd annual meeting of the association for computational linguistics (volume 1: long papers), 2014, pp. 956–966.




\end{thebibliography}

\end{document}